\documentclass[letterpaper]{article} 
\usepackage{aaai2026}  
\usepackage{times}  
\usepackage{helvet}  
\usepackage{courier}  
\usepackage[hyphens]{url}  
\usepackage{graphicx} 
\usepackage{natbib}  
\usepackage{caption} 
\usepackage{algorithm}
\usepackage{algorithmic}
\usepackage{amsfonts} 
\usepackage{amsmath}
\usepackage[algo2e]{algorithm2e} 
\usepackage{colortbl}
\usepackage{graphicx}
\usepackage{multirow}
\usepackage{makecell}
\usepackage{xcolor}
\usepackage{color, soul}
\SetKwComment{Comment}{/* }{ */}

\usepackage{newfloat}
\usepackage{listings}
\DeclareCaptionStyle{ruled}{labelfont=normalfont,labelsep=colon,strut=off} 
\floatstyle{ruled}
\newfloat{listing}{tb}{lst}{}
\floatname{listing}{Listing}
\usepackage[misc]{ifsym}
\newcommand\blfootnote[1]{%
    \begingroup
    \renewcommand\thefootnote{}\footnote{#1}%
    \addtocounter{footnote}{-1}%
    \endgroup
}
\usepackage[hang]{footmisc}

\nocopyright

\title{OmniAvatar: Efficient Audio-Driven Avatar Video Generation \\ with Adaptive Body Animation}
\author{
    Qijun Gan\textsuperscript{\rm 1*},
    Ruizi Yang\textsuperscript{\rm 1*},
    Jianke Zhu\textsuperscript{\rm 1},
    Shaofei Xue\textsuperscript{\rm 2},
    Steven Hoi\textsuperscript{\rm 2}
}

\affiliations{
    \textsuperscript{\rm 1}Zhejiang University, \textsuperscript{\rm 2}Alibaba Group\\
    \{ganqijun,yangruizi,jkzhu\}@zju.edu.cn, \{shaofei.xsf,stevenhoi\}@alibaba-inc.com \\
}

\begin{document}

\twocolumn[{
\renewcommand\twocolumn[1][]{#1}
\maketitle
\begin{center}
    \captionsetup{type=figure}
        \vspace{-0.25in}

    \includegraphics[width=1.0\textwidth]{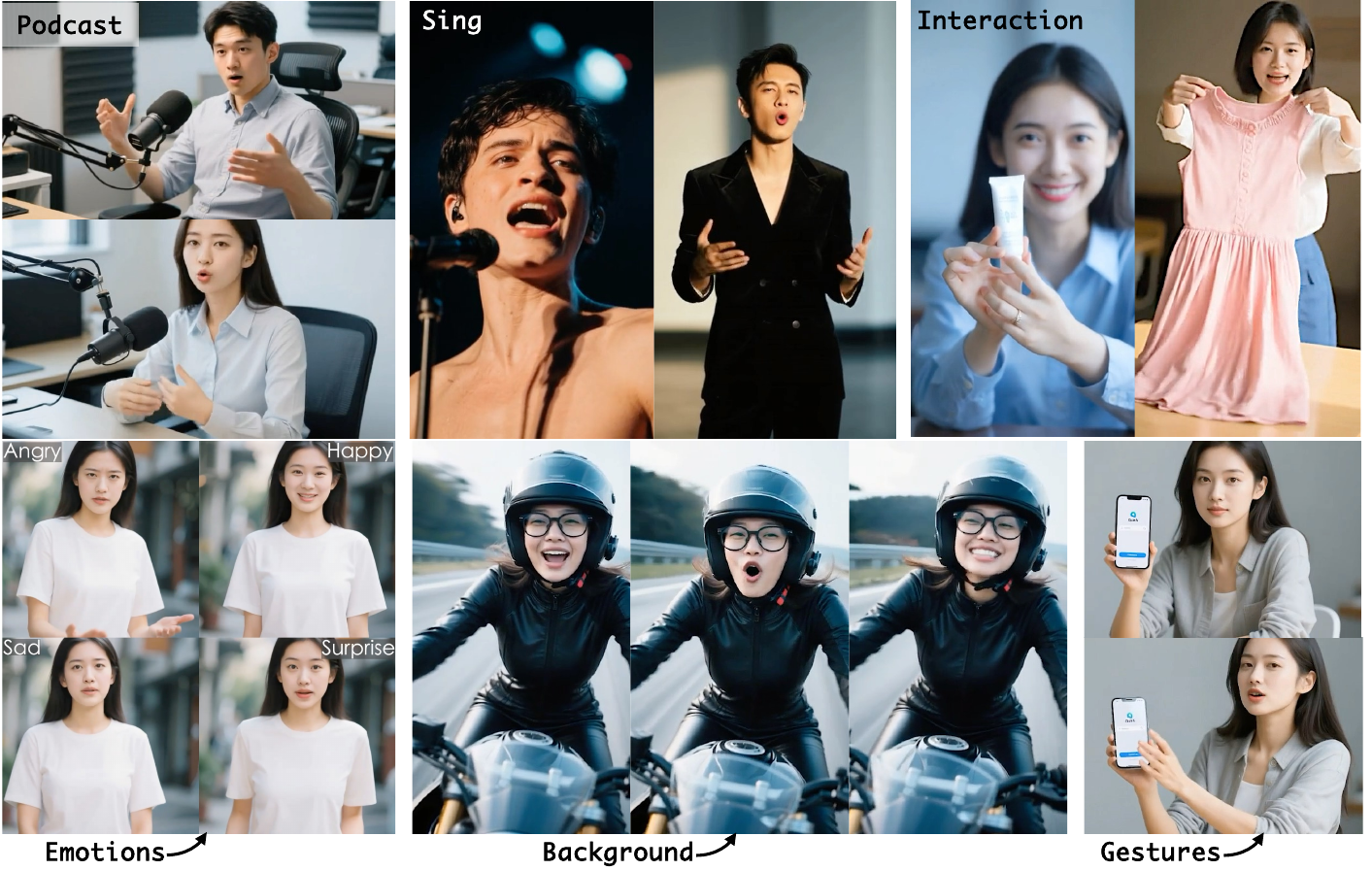}
    \vspace{-0.15in}
    \label{fig:teaser}
    \caption{We introduce OmniAvatar, an innovative framework designed for avatar video generation based on audios and prompts across various scenes. By providing an audio clip and corresponding prompt, OmniAvatar produces a video where lip movements align with the audio, and the scene reflects the prompt.}
\end{center}
}]

{
    \blfootnote{
        \hspace{-0.15in} {\rm *} Work done during internship at Alibaba Group.}
}

\begin{abstract}
Significant progress has been made in audio-driven human animation, while most existing methods focus mainly on facial movements, limiting their ability to create full-body animations with natural synchronization and fluidity. They also struggle with precise prompt control for fine-grained generation. To tackle these challenges, we introduce OmniAvatar, an innovative audio-driven full-body video generation model that enhances human animation with improved lip-sync accuracy and natural movements. OmniAvatar introduces a pixel-wise multi-hierarchical audio embedding strategy to better capture audio features in the latent space, enhancing lip-syncing across diverse scenes. To preserve the capability for prompt-driven control of foundation models while effectively incorporating audio features, we employ a LoRA-based training approach. Extensive experiments show that OmniAvatar surpasses existing models in both facial and semi-body video generation, offering precise text-based control for creating videos in various domains, such as podcasts, human interactions, dynamic scenes, and singing. Our project page is https://omni-avatar.github.io/.
\end{abstract}
\section{Introduction}

The ability to generate realistic and expressive human avatars from conditions has become a cornerstone of digital human research. Portrait video generation, which focuses on generating high-quality visual representations of human faces and bodies from audio or other inputs, plays a critical role in fields such as virtual assistants and film production. As interactive AI and virtual environments become increasingly sophisticated, the need for dynamic, lifelike avatars that can engage users through natural and expressive movements, rather than a talking face, is more important than ever. 

The recent advancements~\cite{meng2024echomimicv2,wei2025mocha,chen2025hunyuanvideoavatar,cui2024hallo3,jiang2024loopy,wang2025fantasytalking,lin2025omnihuman} in audio-driven human animation have significantly improved the realism and naturalness of character generation. However, most existing methods~\cite{ji2024sonic,wang2024v,wei2024aniportrait} focus on driving only the facial movements based on the audio input, limiting their ability to produce full-body animations that have natural movements of the human body. 

To address this challenge, methods such as Hallo3~\cite{cui2024hallo3}, FantasyTalking~\cite{wang2025fantasytalking}, and HunyuanAvatar~\cite{chen2025hunyuanvideoavatar} have adapted current state-of-the-art text-to-video (T2V) or image-to-video (I2V) base models~\cite{yang2024cogvideox,wan2025wan,kong2024hunyuanvideo} for audio-driven animation. Echomimicv2~\cite{meng2024echomimicv2}, Tango~\cite{liu2024tango} and Cyberhost~\cite{lin2024cyberhost} leverage motion-related conditions to generate human animation videos, focusing on the integration of motion data to enhance the generation of realistic body movements. Despite recent progress, methods in full-body animation face several challenges. First, training a full-body model introduces complexities, particularly in maintaining accurate lip-syncing while generating coherent and realistic body movements. Second, current models often struggle with generating natural body movements, leading to stiff or unnatural poses. Moreover, text-based control of body gestures and background movements remains challenging, limiting the customization and adaptability of the generated avatars in dynamic contexts.

To address these challenges, we propose OmniAvatar, a novel model for audio-driven full-body video generation with adaptive body animation. To enhance the naturalness of generated body movements, rather than cross-attention~\cite{cui2024hallo3,wang2025fantasytalking}, the audio features are mapped into the latent space using a proposed audio pack method, and then embedded into the latent space at the pixel level. This approach improves the  ability to perceive and incorporate audio features spatially. To address the lip-syncing challenges across diverse human scenes, we introduce a multi-hierarchical audio embedding strategy, which ensures accurate synchronization between audio and lip movements. Additionally, by utilizing LoRA-based training, we preserve the foundation model’s capabilities while efficiently adapting it to handle the newly introduced audio features. This enables OmniAvatar to produce high-quality videos where the generated avatars not only have accurate lip-syncing but also display realistic and adaptive full-body animations. Meanwhile, OmniAvatar demonstrates greater sensitivity to text conditions and provides more controllable generation.

Extensive experiments show that our model achieves leading results in both facial and semi-body portrait video generation on test datasets. As illustrated in Fig. 1, our model also supports more precise text-based control, making it proficient in generating videos across various domains, including podcasts, human-object interactions, dynamic scenes, and singing. 

In summary, our contributions are as follows:

\begin{itemize}
\item We propose a LoRA-based audio-conditioned portrait video generation model, which enables natural and adaptive body movements and accurate text-based control for generating human animation videos.

\item Our multi-hierarchical pixel-wise audio embedding method improves lip-sync accuracy, ensuring precise synchronization between audio and lip movements.

\item Our model is capable of generating videos featuring natural human body movements, controllable emotions and gestures, and dynamic backgrounds, making it versatile and effective in various applications and scenarios.
\end{itemize}
\section{Related Work}

\subsection{Video Generation}

Recent progress in video generation builds on the success of diffusion-based image synthesis~\cite{dhariwal2021diffusion,rombach2022high}. UNet-based diffusion models pretrained on images are first extended to the temporal domain. Make-A-Video~\cite{singer2022make} inserts temporal attention to create short clips with strong spatial fidelity, while AnimateDiff~\cite{guo2023animatediff} adds motion-aware modules and cross-frame constraints to improve temporal smoothness and enable human-centric content.

Building on these foundations, models shift toward transformer-based architectures to better capture long-range temporal dependencies and semantic consistency. CogVideo~\cite{yang2024cogvideox} and Goku~\cite{chen2025goku} advance text-to-video generation by leveraging large-scale vision-language pretraining and hierarchical token fusion, enabling fine-grained semantic alignment from text prompts. HunyuanVideo~\cite{kong2024hunyuanvideo} extends this line of work by integrating multimodal prompts into a dual-stream DiT-based framework, supporting precise conditional control across diverse scenarios. Wan~\cite{wan2025wan} further contributes a suite of large-scale video generation models with strong scalability and competitive performance.

\subsection{Audio-Driven Video Generation}

Early audio-driven human animation relies on two-stage pipelines that first predict motion parameters—typically via 3D Morphable Models (3DMM)~\cite{blanz2003face}—and then render frames~\cite{zhang2023sadtalker,shen2023difftalk,wei2024aniportrait}. Although interpretable, these cascaded methods suffer from limited expressiveness and temporal drift.

\begin{figure*}[t]
    \centering
    \includegraphics[width=1.0\linewidth]{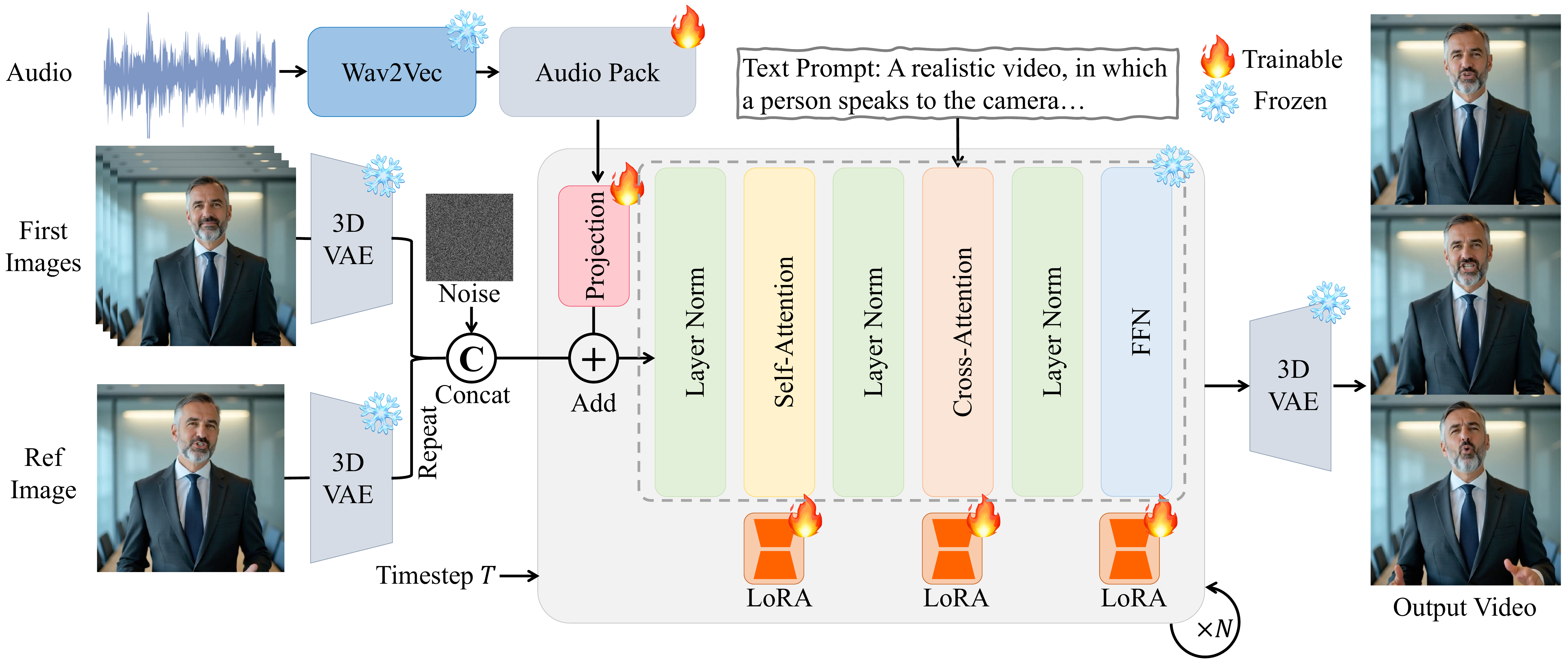}
    \vspace{-0.25in}
    \caption{\textbf{Overview of OmniAvatar.} Our design integrates the simplicity of Wan~\cite{wan2025wan}. Based on text, image, and audio inputs, OmniAvatar generates lifelike human videos, producing highly realistic and expressive character animations.
}
    \vspace{-0.15in}
    \label{fig:framework}
\end{figure*}

Driven by diffusion models, recent research converges on \emph{end-to-end} generation. Representative diffusion-based systems introduce multi-stage refinement and long-range attention to boost realism and coherence~\cite{chen2025echomimic,meng2024echomimicv2,xu2024hallo,cui2024hallo2,cui2024hallo3}. V-Express~\cite{wang2024v} leverages conditional dropout or cyclic prediction to balance weak (audio) and strong (visual) cues. Loopy~\cite{jiang2024loopy} leverage long-term motion information to earn
natural motion patterns. Under data-lean or ambiguous settings, emotion-aware objectives and end-effector guidance refine lip-sync fidelity and facial detail~\cite{tian2024emo,tian2025emo2,wang2025fantasytalking,wei2025mocha,ji2024sonic}. Meanwhile, localized attention and gesture-adaptive conditioning improve controllability and efficiency for semi-body synthesis~\cite{lin2024cyberhost,liu2024tango,guo2024liveportrait}. Toward foundation-level capability, OmniHuman~\cite{lin2025omnihuman} unifies one-stage generation across different identities and scenarios with multiple conditions, whereas HunyuanVideo-Avatar~\cite{chen2025hunyuanvideoavatar} and MultiTalk~\cite{kong2025let} scales to multi-character, high-fidelity human video generation. While current models are still facing challenges in achieving lip-sync accuracy and generating fluid, natural full-body movements simultaneously in audio-driven human video generation.

\section{OmniAvatar}

OmniAvatar aims to create talking avatar videos with the input of a single reference image, audio, and a prompt, while featuring adaptive and natural body animations. The overview of OmniAvatar is illustrated in Fig.~\ref{fig:framework}. To capture audio features at multiple levels,  we introduce a multi-hierarchical audio embedding strategy which maintains pixel-wise alignment between the audio and video (Sec 3.2). Additionally, to retain the powerful capabilities of the foundation model while incorporating audio as new conditions, we apply LoRA-based training to layers of the DiT model (Sec 3.3). To maintain consistency and temporal continuity in long video generation, we leverage frame overlapping and reference image embedding strategy (Sec 3.4).

\subsection{Preliminaries}
\noindent\textbf{Diffusion Models.}
We employ a latent diffusion model (LDM)~\cite{rombach2022high} for efficient video generation, which learns to reverse a diffusion process upon the latent space, progressively transforming noise into data. The process begins with latents $z$ of data (e.g., images, videos) being corrupted by Gaussian $\epsilon$ noise over $t$ steps, $z_t = \sqrt{\alpha_t}z + \sqrt{1-\alpha_t}\epsilon$, where $\alpha_t$ represents the noise scheduler. The model $\epsilon_\theta$ is trained to reverse this noise diffusion process by 
$$
    \mathcal{L} = \mathbb{E}_{t, z_t, c, \epsilon \sim \mathcal{N}(0, 1)} \left[ \| \epsilon_\theta(z_t, t, c) - \epsilon \|^2_2 \right],
$$ where $c$ denotes the conditions. In essence, the model learns to denoise the noisy data step by step to recover the original sample. Diffusion models have demonstrated impressive performance in generating high-quality samples in various domains, including image and video synthesis. Specially, the latent $z$ can be encoded from VAE~\cite{kingma2013auto} encoder and be decoded back to the raw data with VAE decoder.

\noindent\textbf{Diffusion Transformers.}
Diffusion Transformers (DiT)~\cite{peebles2023scalable} extend traditional diffusion models by utilizing transformer architecture~\cite{vaswani2017attention} to model the denoising process. The DiT architecture replaces traditional convolutional layers with self-attention mechanisms, allowing it to learn more complex dependencies in the data, which is particularly useful when handling high-dimensional video or long sequences. Specifically, we adopt Wan2.1~\cite{wan2025wan} as the foundational model. By employing a transformer-based denoising network with full-attention in the latent space, DiT improves the model's capacity to generate high-fidelity and consistent video sequences over long periods, an essential property in avatar video generation.

\noindent\textbf{Low-Rank Adaptation.}
Low-Rank Adaptation (LoRA)~\cite{hu2022lora} improves the efficiency of fine-tuning large models by introducing low-rank decomposition into weight matrices, reducing the number of trainable parameters while retaining the model’s adaptability. This is especially effective for adapting pre-trained models, without requiring full retraining. LoRA achieves this by updating the weight matrices with low-rank approximations during training
$$
W' = W + \Delta W, \quad \Delta W = A B,
$$
where $W$ is the original weight matrix, and $\Delta W$ is the low-rank update, with $A$ and $B$ being the low-rank matrices. This allows the model to efficiently adapt to new tasks, while maintaining high-quality output and low training computational cost.

\subsection{Audio Embedding Strategy}

\begin{figure}
    \centering
    \includegraphics[width=0.5\linewidth]{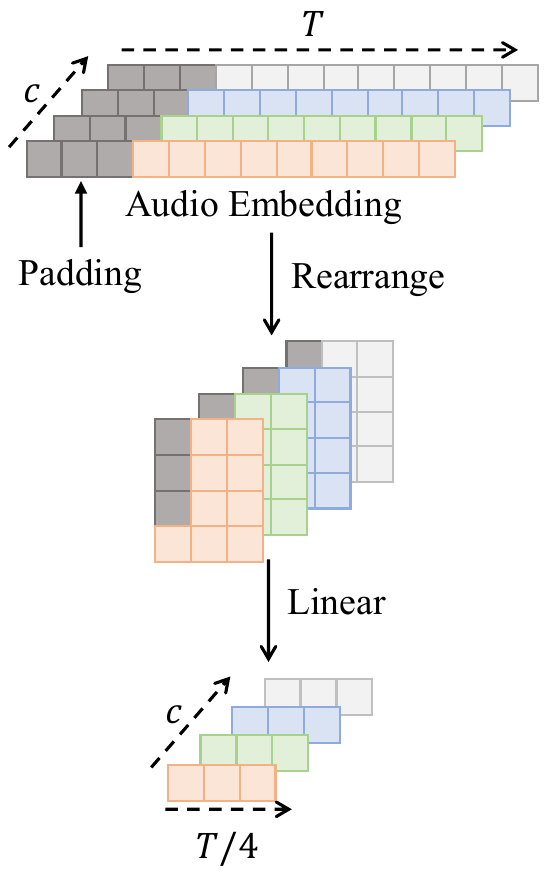}
    \vspace{-0.1in}
    \caption{\textbf{Design of Audio Pack.} To align the audio features to the latent space, audio pack rearranges the padded audio, and then map into audio latent by a linear layer.}
    \vspace{-0.1in}
    \label{fig:audio_pack}
\end{figure}

Most existing methods~\cite{cui2024hallo3,wang2025fantasytalking,chen2025hunyuanvideoavatar,meng2024echomimicv2} typically rely on cross-attention mechanisms to introduce audio features, where the audio information is conditioned on visual features through attention layers. While this approach can lead to good results, it introduces additional computational overhead and tends to overly focus on the relationship between the audio and facial features. In contrast, we propose a pixel-wise audio embedding strategy, where audio features are directly incorporated into the model’s latent space at pixel level. By embedding audio features with pixel-wised fusion, we naturally align lip movements with the audio. And by ensuring the audio information is evenly distributed across the entire video pixels, model results in more holistic and natural body movements in response to the audio.

Given an audio sequence of length $T$, we use Wav2Vec2~\cite{baevski2020wav2vec} for audio feature extraction. Each video, with a length $T$, is compressed into $\frac{T+3}{4}$ latent frames using a pretrained 3D VAE, where the factor of 4 is the time compression ratio of the VAE. To ensure temporal alignment between the audio features and the compressed video latent, we follow the compression pattern of the VAE. First, we pad the audio feature sequence $a$ before the initial frame to match the time length $T+3$. Then, the audio features are grouped into pieces with a compression rate of 4, matching the VAE latent space compression rate, and are subsequently mapped to the latent space $z_a$ with audio pack module. Audio pack compresses the rearranged audio with linear mapping, as shown in Fig.~\ref{fig:audio_pack}. 

To integrate the audio features into the video latent space, we project the audio latent into a space that can be aligned with the video latent. Then, the audio latent is embedded into the video latent at pixel level:
$$
z_a = \text{Pack}(a); z'^i_t = z^i_t +\mathcal{P}^i_a(z_a)
$$ 
where $z^i_t$ is the latent vector of $i$ DiT block corresponding to the video at time step $t$. $\mathcal{P}_a$ denotes the audio projection operation and $\text{Pack}$ refers to the audio compression function. 

By embedding pixel-wise audio features into the video latent space, the generated human motions are adaptively guided by the audio input. To ensure the model effectively learns and retains audio features in deep networks, we employ a multi-hierarchical audio embedding approach that integrates audio embeddings at different stages within the DiT blocks. To prevent the audio features from excessively influencing the latent features, we apply audio embeddings only to the layers between the second and the middle layers of the model. Additionally, the weights for these layers are not shared, allowing the model to maintain separate learning paths for different levels of audio integration.


\subsection{LoRA-based DiT Optimization}
Previous methods for audio-conditioned diffusion models typically follow one of two strategies: either training the full model~\cite{lin2025omnihuman,chen2025hunyuanvideoavatar} or fine-tuning only specific layers~\cite{cui2024hallo3, wang2025fantasytalking}. When performing full training, we notice that updating all layers leads to degradation in the capability of the model to generate coherent and high-quality video sequences. Specifically, the model generates unrealistic or static content more easily, while struggling to capture fine details. This occurs because the model overfits to the human speech datasets, leading to poor generalization and difficulty in controlling the video generation. On the other hand, freezing the DiT model and only fine-tuning the layers responsible for processing the audio features results in poor alignment between audio and video. The lip-sync performance is compromised because the model struggles to accurately map audio features to realistic facial movements.

To overcome these challenges, we propose a balanced fine-tuning strategy based on LoRA. Instead of fine-tuning all layers or just updating the audio-related layers, we use LoRA strategy to adapt the model efficiently. LoRA introduces low-rank matrices into the weight updates of the attention and feed-forward (FFN) layers, allowing the model to learn audio-conditioned behavior without altering the underlying model's capacity. 

\subsection{Long Video Generation}

Generating long, continuous videos is crucial for audio-driven avatar video generation. The ability to generate extended videos, without compromising the visual quality or temporal consistency, presents a significant challenge. To address this, we employ reference image embedding strategy to preserve the identity and frame overlapping for temporal consistency. Algo.~\ref{algo:infer} shows the inference pipeline for long video generation.

\begin{algorithm}
\caption{Long Video Inference}
\KwIn{Audio latents $z_a$ with length $l$ , Pretrained model, Inference length $s$, Overlap length $f$, First frame latents $z_{\text{ref}}$}
\KwOut{Denoised video latents $z_0$}
\SetKwFunction{FMain}{LongVideoInference}
\SetKwProg{Fn}{Function}{:}{}
\Fn{\FMain{$a$, $l$, $s$, $f$, $z_{\text{ref}}$}}{
    \(N, l_{\text{pad}} = \text{FindLoopN}(l,s)\)           \tcp*{get loop times}
    
    \(z_a \leftarrow \text{Zeros}(1) + z_a + \text{Zeros}(l_{\text{pad}})\)\tcp*{pad input}
    
    \(z_T \leftarrow z_{\text{ref}} + \text{Noise}(l + l_{\text{pad}})\);
    
    \(l = l + l_{\text{pad}} + 1;n = 0\);
    
    \For{$i = 1, \dots, N$}{

        \If{$i = 0$}{   
        \tcp{use first frame as prefix}
            \( f_i = 0 \);
            \( z_{\text{prefix}} = z_{\text{ref}}\);
        }
        \Else{        
        \tcp{use previous suffix}
            \(f_i = f \);
            \(z_T \leftarrow z_{\text{prefix}} + z_T\);
            }
        \(n = n - f_i\); \tcp*{overlap}
        
        \( z_0^{[n, n+s]} = \text{Model}(z_a^{[n, n+s]}, z_T^{[0, s]}, z_{\text{ref}}) \); 
        
        \(z_T = z_T^{[s, l]}\);
        \(z_{\text{prefix}} = z_0^{[n+s-f, n+s]}\);
        
        \(n = n + s\); \tcp*{move to the next clip}
        }
    \Return{Denoised latent $z_0$}
}
\label{algo:infer}
\end{algorithm}

\noindent\textbf{Identity Preservation.}
To preserve the identity throughout the video generation process, we utilize reference image embedding strategy, which introduce a reference frame that serves as a fixed guidance of identity. Specifically, we extract the latent representation of the reference frame and repeat it to match the length of the video. This repeated reference latent is then concatenated with the video latent at each time step, ensuring that the avatar's appearance remains consistent across all frames. By using this reference frame, we effectively anchor the identity of the avatar, ensuring its visual characteristics remain consistent throughout the video sequence.

\noindent\textbf{Temporal Consistency.}
Maintaining temporal consistency is crucial for creating long, continuous videos with smoothing frame transitions. To achieve seamless video continuity, we use a latent overlapping strategy. We train the model with a combination of single-frame and multi-frame prefix latents. During inference, the first batch of frames is generated using the reference frame as both the prefix latent and identity guidance. For subsequent batches, the last frames from the previous batch serve as the prefix latents, while the reference frame remains fixed to guide identity. This overlap ensures smooth transitions between video segments, preserving temporal continuity and preventing abrupt changes in motion or appearance.

\section{Experiments}
\begin{table*}[htp!]
\centering
\small
\begin{tabular}{c|cccccc}
\Xhline{1pt}
\textbf{Methods} & \textbf{FID}   $\downarrow$ & \textbf{FVD}   $\downarrow$  & \textbf{Sync-C}  $\uparrow$  & \textbf{Sync-D}    $\downarrow$  & \textbf{IQA}  $\uparrow$  & \textbf{ASE}  $\uparrow$ \\ \hline
\multicolumn{7}{c}{HDTF} \\ \hline
Sadtalker~\cite{zhang2023sadtalker}  & 50.0 & 538 & 7.01 & 8.54 & 3.16 & 2.23 \\ 
Aniportrait~\cite{wei2024aniportrait} & 46.1* & 546* & 3.64* & 10.79* & 3.96* &  2.35* \\ 
V-express~\cite{wang2024v} & 59.1* & 548* & 8.02* & 7.69* &  3.32* & 1.96* \\ 
EchoMimic~\cite{chen2025echomimic} & 61.7* & 575* & 5.71* & 9.14* & 3.61* & 2.19*   \\ 
Hallo3~\cite{cui2024hallo3} & 42.1 & 406 & 6.89 & 8.71 & 3.55 & 2.15 \\
FantasyTalking~\cite{wang2025fantasytalking} & 43.9 & 441 & 3.75 & 11.0 & 3.59 & 2.17   \\ 
HunyuanAvatar~\cite{chen2025hunyuanvideoavatar} & 47.3 & 588 & 7.31 & 8.33 & 3.58 & 2.20 \\ 
MultiTalk~\cite{kong2025let} & 44.2 & 436 & 7.63 & 7.78 & 3.54  & 2.14  \\ 
GT & - & - & 8.20 & 6.89 & 3.94  &  2.48 \\ 
\cellcolor[HTML]{D9D9D9}{Ours} & \cellcolor[HTML]{D9D9D9}{\textbf{37.3}} & \cellcolor[HTML]{D9D9D9}{\textbf{382}} & \cellcolor[HTML]{D9D9D9}{\underline{7.62}} & \cellcolor[HTML]{D9D9D9}{\underline{8.14}} & \cellcolor[HTML]{D9D9D9}{\textbf{3.82}} & \cellcolor[HTML]{D9D9D9}{\textbf{2.41}} \\ \Xhline{1pt}
  \multicolumn{7}{c}{AVSpeech-Face}  \\  \hline
Sadtalker~\cite{zhang2023sadtalker}  & 103 & 1182 & 4.31 & 9.68 & 2.45 & 1.49 \\ 
Aniportrait~\cite{wei2024aniportrait} & 100 & 1095 & 2.09 & 11.6 & 2.31 & 1.39 \\ 
V-express~\cite{wang2024v} & 194 & 1589 & 6.19 &  8.88  & 2.26 & 1.54 \\ 
EchoMimic~\cite{chen2025echomimic} & 69.1 & 751 & 4.31 & 9.81 & 2.71 & 1.56 \\ 
Hallo3~\cite{cui2024hallo3} & 68.6 & 703 & 4.93 & 9.76 & 2.64 & 1.49 \\ 
FantasyTalking~\cite{wang2025fantasytalking} & 86.1 & 885 & 3.34 & 11.1 & 2.69 & 1.57  \\ 
HunyuanAvatar~\cite{chen2025hunyuanvideoavatar} & 88.6 & 796 & 5.97 & 8.66 & 2.52 & 1.45\\ 
MultiTalk~\cite{kong2025let} & 78.3 & 729 & 6.23 & 8.43 & 2.74  &  1.59 \\ 
GT  & - & - & 7.08 & 8.07 & 2.54 & 1.47 \\
\cellcolor[HTML]{D9D9D9}{Ours} & \cellcolor[HTML]{D9D9D9}{\textbf{66.5}} & \cellcolor[HTML]{D9D9D9}{\textbf{692}} & \cellcolor[HTML]{D9D9D9}{\textbf{6.32}} & \cellcolor[HTML]{D9D9D9}{\textbf{8.38}} & \cellcolor[HTML]{D9D9D9}{2.63} & \cellcolor[HTML]{D9D9D9}{1.51} \\ \Xhline{1pt}
\end{tabular}
\vspace{-0.1in}
\caption{Quantitative comparison on the test set with existing audio-driven talking face video generation methods.}
\vspace{-0.3in}
\label{tab:face}
\end{table*}

\begin{table*}[htp!]
\centering
\small
\begin{tabular}{c|cccccc}
\Xhline{1pt}
\textbf{Methods} &  \textbf{FID}   $\downarrow$ & \textbf{FVD}   $\downarrow$  & \textbf{Sync-C}  $\uparrow$  & \textbf{Sync-D}    $\downarrow$  & \textbf{IQA}  $\uparrow$  & \textbf{ASE}  $\uparrow$ \\ \hline
Hallo3~\cite{cui2024hallo3} & 104 & 1078 & 5.23 & 9.54 & 3.41 & 2.00 \\
FantasyTalking~\cite{wang2025fantasytalking} & 78.9 & 780 & 3.14 & 11.2 & 3.33 & 1.96   \\ 
HunyuanAvatar~\cite{chen2025hunyuanvideoavatar} & 77.7 & 887 & 6.71 & 8.35 &  3.61 & 2.16 \\ 
MultiTalk~\cite{kong2025let} & 74.7 & 787 & 4.76 & 9.99 &  3.67 & 2.22  \\ 
GT & - & - & 6.75 & 7.76 & 3.92  &  2.38 \\ 
\cellcolor[HTML]{D9D9D9}{Ours} & \cellcolor[HTML]{D9D9D9}{\textbf{67.6}} & \cellcolor[HTML]{D9D9D9}{\textbf{664}} & \cellcolor[HTML]{D9D9D9}{\textbf{7.12}} & \cellcolor[HTML]{D9D9D9}{\textbf{8.05}} & \cellcolor[HTML]{D9D9D9}{\textbf{3.75}} & \cellcolor[HTML]{D9D9D9}{\textbf{2.25}} \\ \Xhline{1pt}
\end{tabular}
\vspace{-0.1in}
\caption{Quantitative comparison on the AVSpeech~\cite{ephrat2018looking} test set with existing audio-driven semi-body video generation methods. $*$ denotes the test videos maybe are used to train by the methods.}
\vspace{-0.3in}
\label{tab:semi-body}
\end{table*}

\begin{figure*}
    \centering
    \includegraphics[width=0.95\linewidth]{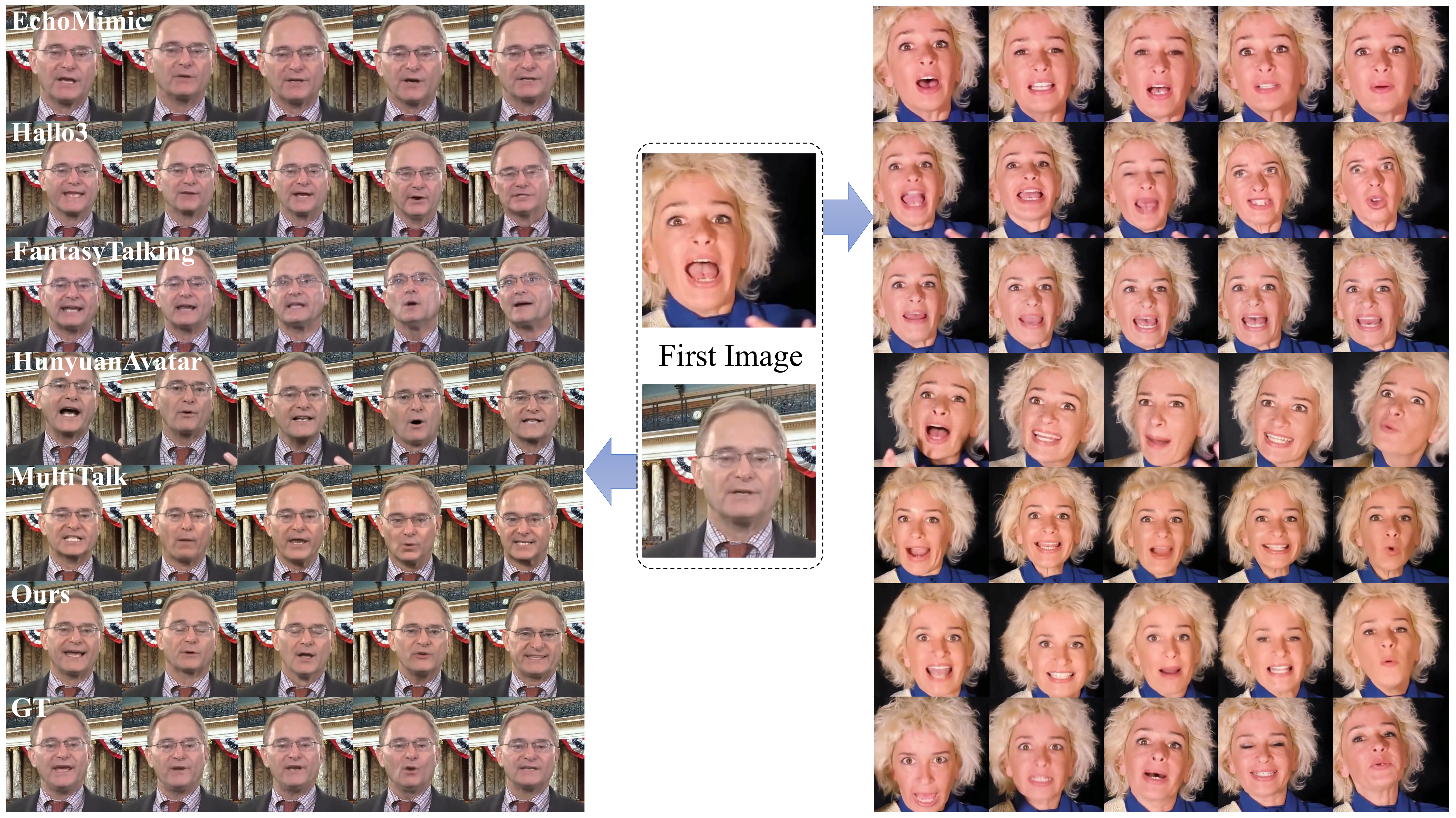}
    \vspace{-0.1in}
    \caption{The qualitative comparison on HDTF~\cite{zhang2021flow} and AVSpeech~\cite{ephrat2018looking} for facial generation. We use cropped square faces as input in the AVSpeech test set.}
    \vspace{-0.1in}
    \label{fig:face}
\end{figure*}
\subsection{Experimental Setups}

\noindent\textbf{Implementation.}
To train OmniAvatar, we use Wan2.1-T2V-14B~\cite{wan2025wan} as the base model. The training process consists of two phases. In the first phase, we train the model on low-resolution (480p) audio-video data to establish fundamental audio-visual alignment. In the second phase, we combine both low-resolution and high-resolution audio-video data to further refine the model, with the goal of improving motion stability.
The maximum latent token length for video during training is set to 30,000, and 10\% of the data is dropped for audio to perform classifier-free guidance. During training, we pre-extract the video latents and caption embeddings for efficiency and randomly select the length of prefix latent between 1 and 4. For the LoRA optimization, we set the rank to 128 and the alpha to 64, ensuring a balance between efficient fine-tuning and preserving the performance of the base model.
The training process is conducted on 64 A100 80GB GPUs, with a learning rate set to 5e-5.

During inference, we apply a 13-frame video overlap for long video generation. The denoising process runs for 25 steps, with both the audio and text classifier-free guidance (CFG) set to 4.5 for stable video generation.

\begin{figure*}
    \centering
    \includegraphics[width=1.0\linewidth]{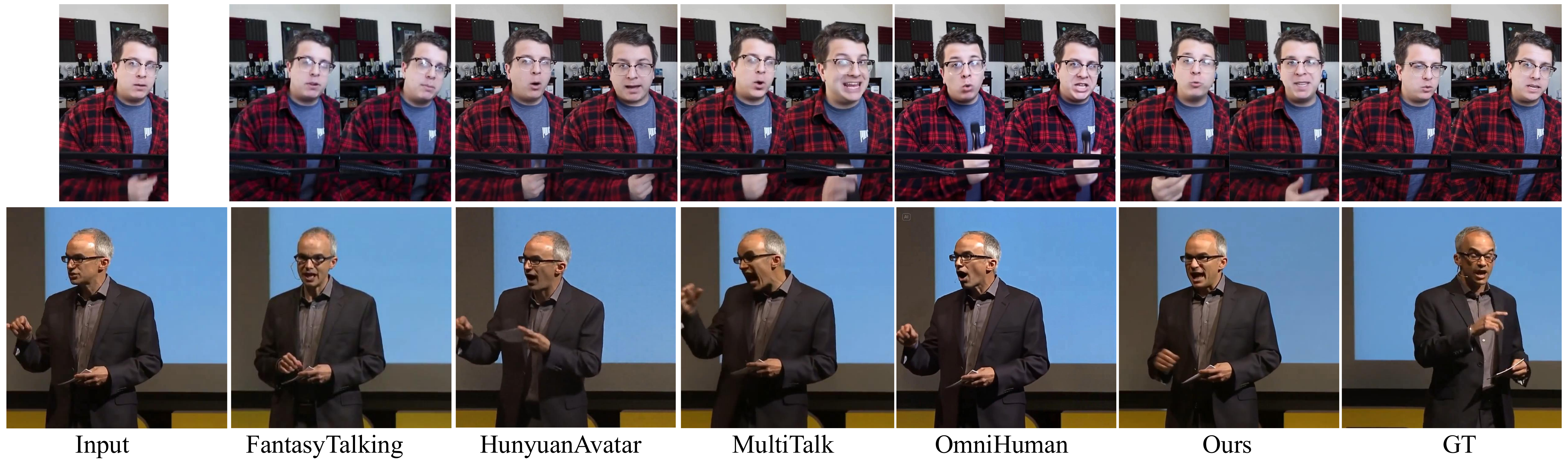}
    \vspace{-0.25in}
    \caption{Visual comparison for semi-body generation on AVSpeech~\cite{ephrat2018looking} test set.}
    \vspace{-0.1in}
    \label{fig:fullbody}
\end{figure*}

\noindent\textbf{Dataset.}
We use the fully open-source AVSpeech~\cite{ephrat2018looking} dataset for training our model. AVSpeech is a large-scale audio-visual dataset containing over 4700 hours of human video. To ensure high-quality training data, we utilize SyncNet~\cite{chung2017out} and Q-Align~\cite{wu2023q} to filter for higher-quality videos by evaluating lip-sync accuracy and video fidelity.
After applying these filters, we obtain a subset of 774,207 samples with durations ranging from 3 to 20 seconds, totaling approximately 1,320 hours of data. From this dataset, we randomly select 100 samples for the semi-body test set, with the remaining data used for training. For comprehensive evaluation with existing methods, we select 100 samples from the HDTF~\cite{zhang2021flow} dataset as an extra talking face test set.

\noindent\textbf{Metrics.}
To validate the performance of our model, we use FID~\cite{heusel2017gans} to evaluate the quality of generated images. For video quality assessment, we use FVD~\cite{unterthiner2018towards}. Additionally, we employ the Q-align~\cite{wu2023q} visual language model to evaluate the video quality (IQA) and aesthetic metrics (ASE). For assessing the synchronization of generated lip movements with audio, we use Sync-C and Sync-D metrics~\cite{chung2017out}.

\subsection{Comparisons with Existing Methods}

\noindent\textbf{Comparison on Talking Face.} We compare OmniAvatar with several existing talking face methods, including Sadtalker~\cite{zhang2023sadtalker}, Aniportrait~\cite{wei2024aniportrait}, V-express~\cite{wang2024v}, EchoMimic~\cite{chen2025echomimic}, Hallo3~\cite{cui2024hallo3}, FantasyTalking~\cite{wang2025fantasytalking},  HunyuanAvatar~\cite{chen2025hunyuanvideoavatar} and MultiTalk~\cite{kong2025let}. The experiments are conducted on two test sets: the HDTF~\cite{zhang2021flow} test set and the cropped face test set from AVSpeech~\cite{ephrat2018looking}. Fig.~\ref{fig:face} presents the qualitative results, demonstrating that our model generates videos with superior image quality, more natural and expressive facial movements, and enhanced visual aesthetics. With our designed pixel-wised audio embedding strategy, the generated videos demonstrate more accurate lip-syncing, and a more realistic alignment between the audio and facial expressions. 

The quantitative results in Tab.~\ref{tab:face} further confirm the superiority of OmniAvatar. Our model achieves leading performance in Sync-C, showcasing superior lip-sync accuracy, which is a key measure for talking face methods. We also achieve competitive results in other metrics like FID, FVD, and IQA, reflecting our model’s ability to generate high-quality and perceptually accurate images and videos. While methods like Sadtalker and V-express achieve decent lip-sync scores, OmniAvatar stands out due to its balance of high video quality, lip-sync precision, and aesthetic appeal. FantasyTalking~\cite{wang2025fantasytalking} and MultiTalk~\cite{kong2025let}, although trained based on Wan~\cite{wan2025wan}, freezes the weights of diffusion blocks, which restricts its ability to align audio and video effectively. HunyuanAvatar~\cite{chen2025hunyuanvideoavatar} demonstrates competitive lip-sync capabilities, while our method achieves superior image quality, offering more visually appealing results while maintaining high lip-sync accuracy.

\noindent\textbf{Comparison on Semi-Body Animation.}
We compare OmniAvatar with FantasyTalking~\cite{wang2025fantasytalking},  HunyuanAvatar~\cite{chen2025hunyuanvideoavatar}, MultiTalk~\cite{kong2025let} and OmniHuman~\cite{lin2025omnihuman} on semi-body animation tasks using the AVSpeech test set in semi-body scenarios. Our method outperforms these existing methods across both qualitative and quantitative evaluations. The qualitative results, shown in Fig.~\ref{fig:fullbody}, demonstrate that OmniAvatar generates more natural and fluid body movements while preserving realistic and synchronized lip-syncing. The generated videos exhibit smoother transitions, more coherent upper-body movements.

Table~\ref{tab:semi-body} shows that OmniAvatar excels in several key metrics for semi-body video generation, especially in audio-lip synchronization and overall video quality. The results confirm that OmniAvatar not only excels at generating realistic body movements but also maintains seamless audio-visual synchronization.

\begin{figure*}
    \centering
    \includegraphics[width=1.0\linewidth]{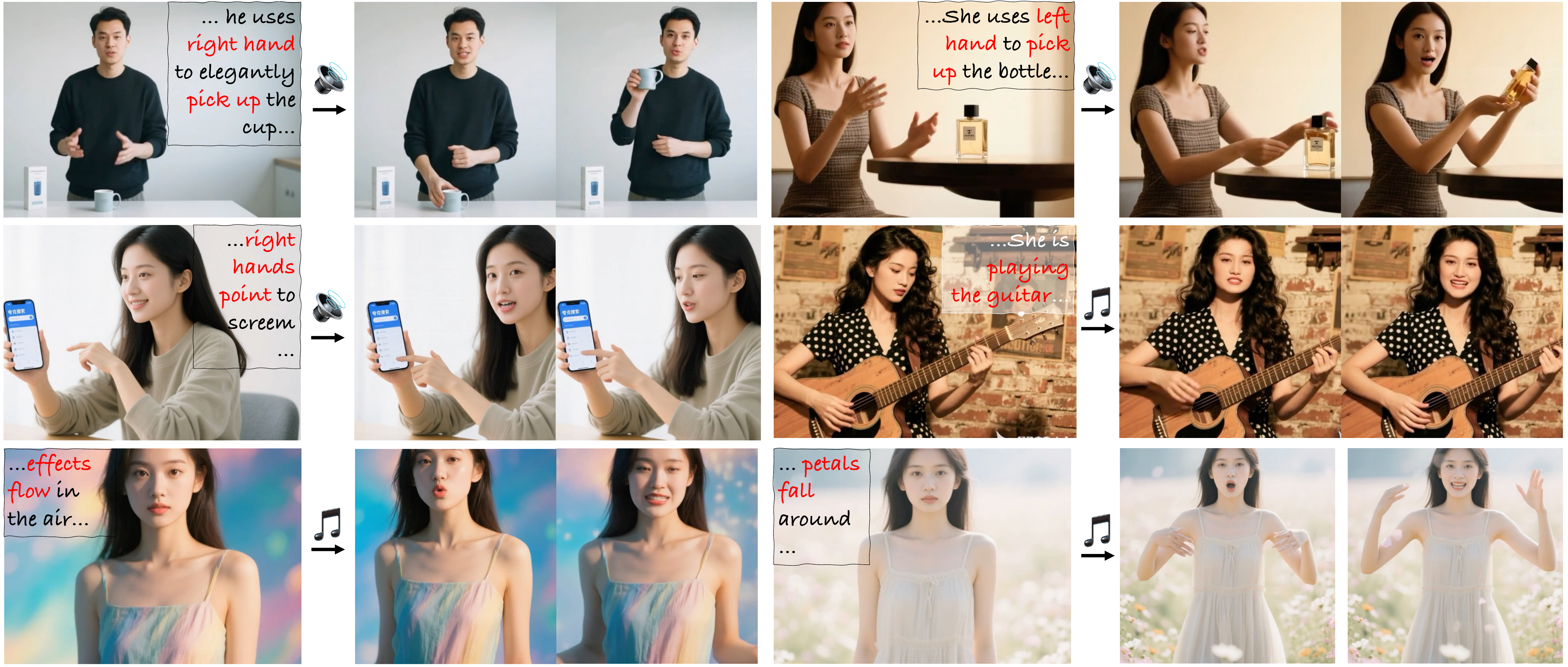}
    \vspace{-0.15in}
    \caption{Driven by a piece of audio and a specific prompt, OmniAvatar can manage various scenes, including human-object interactions, gesture control, and dynamic background configuring.}
    \vspace{-0.15in}
    \label{fig:more1}
\end{figure*}

\begin{figure}
    \centering
    \includegraphics[width=1.0\linewidth]{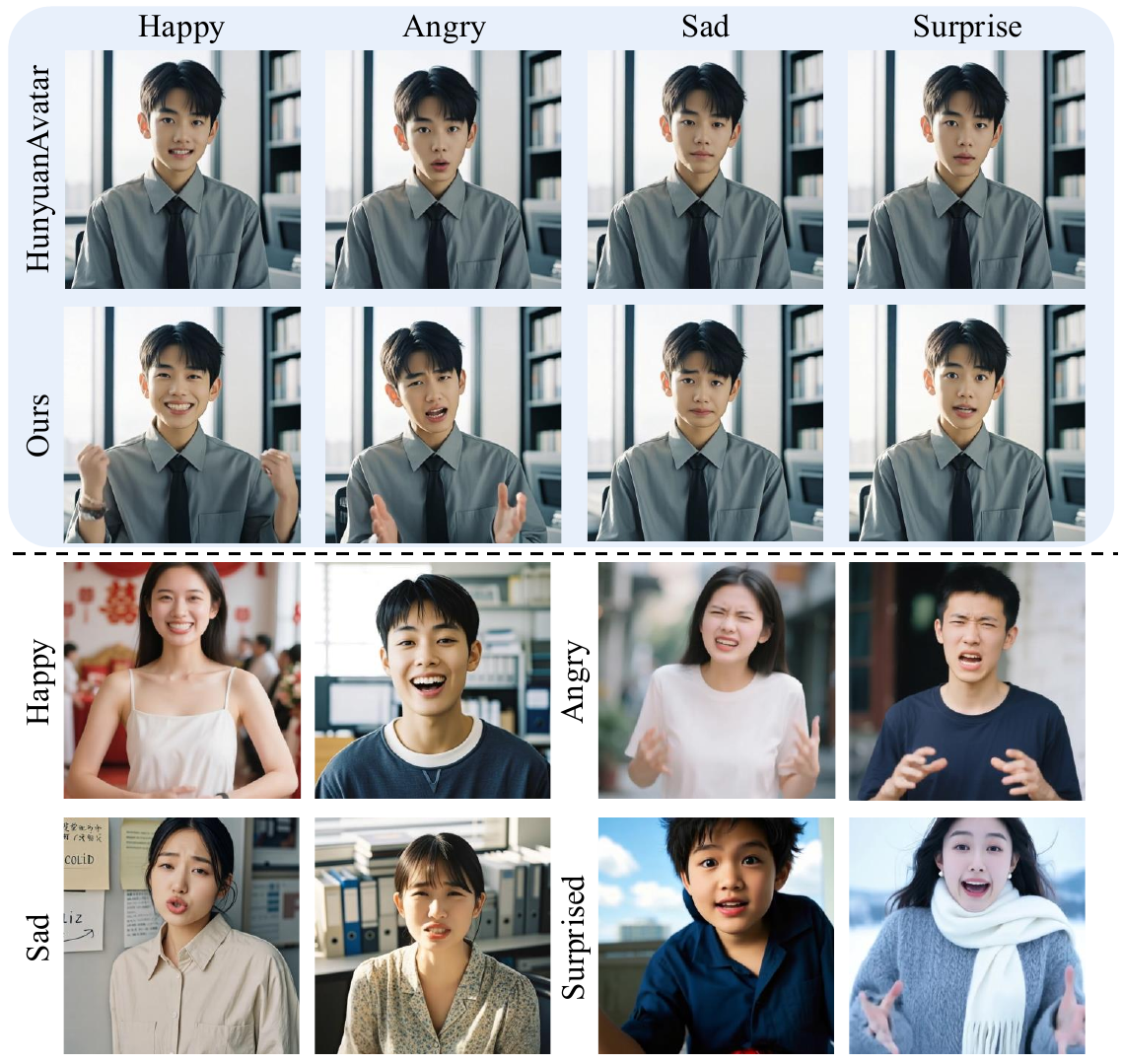}
    \vspace{-0.2in}
    \caption{\textbf{Facial expression control.} By configuring prompts to control the emotions of characters. The two lines above are a comparison with HunyuanAvatar~\cite{chen2025hunyuanvideoavatar}, using the same set of audio and images.}
    \vspace{-0.1in}
    \label{fig:emo}
\end{figure}
\noindent\textbf{More results} The results shown in Fig.~\ref{fig:more1} illustrate the versatility of OmniAvatar in generating realistic video animations. Attributed to the text control capability of Wan2.1-T2V and our LoRA training design, not only can it generate natural human movements in response to audio, but it also supports the interaction between the avatars and surrounding objects. OmniAvatar also enables gesture manipulation and background control, making it a powerful tool for dynamic, interactive, audio-driven video generation across various scenarios. The emotions of characters can also be controlled by prompts, as shown in the Fig.~\ref{fig:emo}.

\subsection{Ablation Studies}

\begin{figure}
    \centering
    \includegraphics[width=1.0\linewidth]{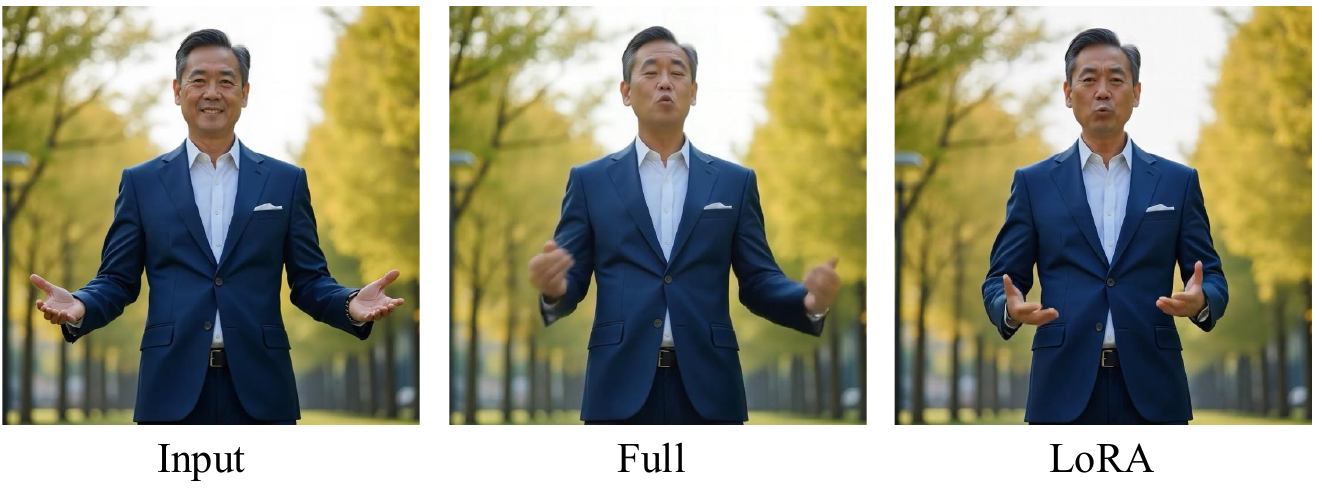}
    \vspace{-0.2in}
    \caption{Ablation study on different training strategies: full training vs. LoRA-based training.}
    \vspace{-0.2in}
    \label{fig:lora}
\end{figure}

\begin{table}[t]
\centering
\begin{tabular}{c|cccc}
\Xhline{1pt}
\textbf{Methods} &  \textbf{FVD}   $\downarrow$  & \textbf{Sync-C}  $\uparrow$  & \textbf{Sync-D}    $\downarrow$  & \textbf{IQA}  $\uparrow$   \\ \hline
Frozen-DiT & 678 & 4.26 & 9.98 & 3.74  \\
Full-Training & 715 & 5.58 & 9.23 & 3.54  \\
LoRA$+$SHE & 685 & 6.58 & 8.73 & 3.73    \\ \hline
CFG-3 & 677 & 6.92 & 8.08 &  3.70 \\ 
CFG-6 & 669 & \textbf{7.37} & \textbf{7.94} & 3.67 \\ \hline
Wan-1.3B & 711 & 3.75 & 9.72 & 2.24 \\ 
\cellcolor[HTML]{D9D9D9}{Ours} & \cellcolor[HTML]{D9D9D9}{\textbf{664}} & \cellcolor[HTML]{D9D9D9}{7.13} & \cellcolor[HTML]{D9D9D9}{8.05} & \cellcolor[HTML]{D9D9D9}{\textbf{3.75}} \\ \Xhline{1pt}
\end{tabular}
    \vspace{-0.1in}
\caption{Ablation study on model design. We conduct experiments on model selection, classifier-free guidance settings and model size. SHE represents audio input through single hierarchical embedding.}
\vspace{-0.1in}
\label{tab:ablation}
\end{table}

\noindent\textbf{Ablation on LoRA and full training.}
Although full training results in faster convergence and better scene adaptation, as shown in Tab.~\ref{tab:ablation}, it may reduce the quality of video generation. Due to the constraints of the dataset quality, particularly the lack of high-resolution portrait data, full training can lead to a degradation in image quality and cause distortions. Motion blur in low-quality data can negatively impact the performance of human motion elements, like hands and mouths, in the generated video, resulting in lower lip-sync scores. As shown in Fig.~\ref{fig:lora}, full training will damage to character details such as hands and eyes. On the other hand, the design of LoRA effectively preserves the original capabilities of the model while seamlessly integrating the newly introduced audio features, allowing for high-quality outputs with the incorporation of additional audio conditioning.

\noindent\textbf{Ablation on Multi- and Single-hierarchical Audio Embedding.} We conduct an ablation experiment by adding audio embedding to only a single layer for comparison. To align with the perception field of the model, the audio embedding is applied at the middle layer. As shown in the Tab.~\ref{tab:ablation}, the multi-hierarchical audio embedding approach leads to better audio synchronization performance. This highlights the benefit of integrating audio features at various levels, enabling more precise alignment between the audio and the generated video.

\noindent\textbf{Ablation on Classifier-Free Guidance (CFG).} The experiment demonstrates that higher values of classifier-free guidance (CFG) improve the synchronization between lip movements and pose generation, resulting in more accurate alignment with the audio. However, excessive high CFG values can lead to exaggerated lip movements, causing unrealistic character expressions and unnatural video generation. Therefore, we choose 4.5 as a reasonable value for CFG of audio and text.
\section{Conclusion}

We propose OmniAvatar, a novel model for audio-driven full-body video generation that improves the naturalness and expressiveness of generated human avatars. By introducing a pixel-wise multi-hierarchical audio embedding strategy and leveraging LoRA-based training, our model addresses the key challenges of synchronizing lip movements and generating realistic, dynamic body movements simultaneously. Extensive experiments on test datasets demonstrate that OmniAvatar achieves state-of-the-art results in both facial and semi-body portrait video generation. Furthermore, our model excels in precise text-based control, enabling the generation of high-quality videos across various domains.
\section{Appendix}

\subsection{More Visualization Results}

\begin{figure*}
    \centering
    \includegraphics[width=1.0\linewidth]{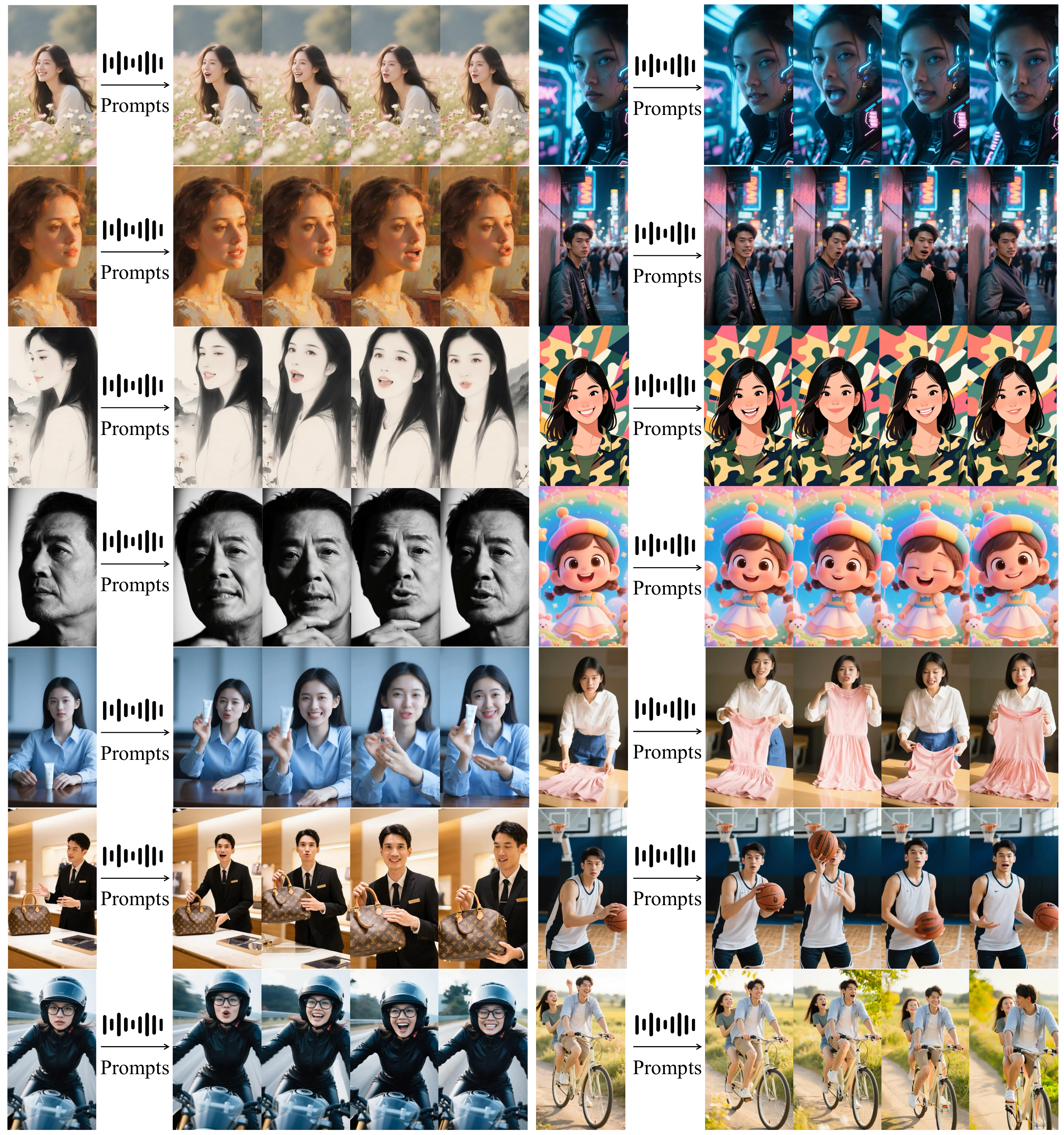}
    \vspace{-0.1in}
    \caption{More visualization results.}
    \vspace{-0.1in}
    \label{fig:more_vis}
\end{figure*}

Fig.~\ref{fig:more_vis} shows the visualization results on more scenarios, such as video with realistic painting style, plain painting style, oil painting style, cartoon painting style, human-object interaction, background moving, etc.

\subsection{Limitation and Discussion}

Despite the significant advances made by OmniAvatar, there are a few limitations. First, our model inherits the weaknesses of the base model, Wan~\cite{wan2025wan}, such as color shifts and error propagation in long video generation. These issues arise as inaccuracies accumulate over time, particularly in extended videos. Second, while LoRA preserves the model’s capabilities, complex text-based control, such as distinguishing which character is speaking or handling multi-character interactions, remains a challenge. 

Additionally, diffusion-based inference requires a large number of denoising steps, resulting in long inference times. This makes real-time video generation challenging, limiting the applicability of the model in scenarios that demand fast, interactive responses. Addressing these issues in future work will improve the efficiency and versatility of OmniAvatar for a wider range of applications.

\bibliography{aaai2026}

\end{document}